\documentclass[lettersize,journal]{IEEEtran}
\usepackage{amsmath,amsfonts}
\usepackage{algorithmic}
\usepackage{algorithm}
\usepackage{array}
\usepackage[caption=false,font=normalsize,labelfont=sf,textfont=sf]{subfig}
\usepackage{textcomp}
\usepackage{stfloats}
\usepackage{url}
\usepackage{verbatim}
\usepackage{graphicx}
\usepackage{cite}
\usepackage[dvipsnames]{xcolor}
\definecolor{aliceblue}{rgb}{0.94, 0.97, 1.0}
\usepackage{circledsteps}
\usepackage{graphicx}
\usepackage{float}
\usepackage{stfloats}
\usepackage{hyperref}
\usepackage{tabularx}
\usepackage{multirow}
\usepackage{booktabs}
\usepackage{amssymb}
\usepackage{bbding}
\usepackage{pifont}
\usepackage{wasysym}
\usepackage{utfsym}
\usepackage{fontawesome}
\usepackage{soul} 
\usepackage{color, xcolor} 
\usepackage{colortbl} 
\usepackage{xcolor}
\usepackage{array}  
\usepackage[normalem]{ulem}

\definecolor{red}{RGB}{255, 0, 0}
\definecolor{blue}{RGB}{0, 0, 255} 
\definecolor{SeaGreen}{RGB}{32, 158, 133} 
\definecolor{LightSeaGreen}{RGB}{32,178,170}
\definecolor{apricot}{rgb}{0.98, 0.81, 0.69}
\definecolor{babyblue}{rgb}{0.54, 0.81, 0.94}
\definecolor{celadon}{rgb}{0.67, 0.88, 0.69}
\definecolor{dollarbill}{rgb}{0.52, 0.73, 0.4}
\definecolor{inchworm}{rgb}{0.7, 0.93, 0.36}
\definecolor{lavenderblue}{rgb}{0.8, 0.8, 1.0}
\definecolor{lightblue}{rgb}{0.68, 0.85, 0.9}
\definecolor{lightgray}{rgb}{0.83, 0.83, 0.83}
\definecolor{grannysmithapple}{rgb}{0.66, 0.89, 0.63}

\begin{document}

\title{DRMOT: A Dataset and Framework for \\ RGBD Referring Multi-Object Tracking}

\author{Sijia Chen$^{1,\ast}$, Shaofeng Liang$^{2,\ast}$, Lijuan Ma$^{3,\ast}$, Yanqiu Yu$^{1}$, Nan Li$^{2}$, Wenshuo Chen$^{2}$,\\ Runwei Guan$^{2}$, Dongming Wu$^{4}$, En Yu$^{1}$, Yutao Yue$^{2,5,\dagger}$\\
\thanks{$^{\ast}$ Equal contribution.}
\thanks{$^{\dagger}$ Corresponding author: yutaoyue@hkust-gz.edu.cn}
\IEEEauthorblockA{$^{1}$Huazhong University of Science and Technology}\\
\IEEEauthorblockA{$^{2}$The Hong Kong University of Science and Technology (Guangzhou)}\\
\IEEEauthorblockA{$^{3}$South-Central Minzu University}
\IEEEauthorblockA{$^{4}$The Chinese University of Hong Kong} \\
\IEEEauthorblockA{$^{5}$Institute of Deep Perception Technology, JITRI}
		\\
		\vspace{-1.2cm}
}

\maketitle

\begin{abstract}
Referring Multi-Object Tracking (RMOT) aims to track specific targets based on language descriptions, playing a vital role in interactive AI systems such as robotics and autonomous driving. However, existing RMOT methods rely exclusively on 2D RGB observations. Since similar 2D projections can correspond to different 3D configurations, the available observations may not uniquely determine the referred targets. This creates a fundamental mismatch between the task requirements and the available input. To advance research on the DRMOT task, we construct \textbf{DRSet}, a dataset tailored for the DRMOT task. DRSet comprises  2,076 referring expressions over 186 RGB-D source sequences. It features synchronized RGB images and depth maps paired with language descriptions, including explicit depth-dependent descriptions, to systematically evaluate spatial-semantic detection and tracking capabilities. Furthermore, we propose \textbf{DRTrack}, an MLLM-guided depth-referring tracking framework. DRTrack leverages a Geometric-Aware GRPO fine-tuned Multimodal Large Language Model (MLLM) to achieve precise spatial-semantic detection from joint L+RGB+D inputs, while incorporating explicit depth cues to enforce robust trajectory association. Extensive experiments on the DRSet dataset demonstrate the effectiveness of our framework. Our dataset and code will be available at \url{https://github.com/chen-si-jia/DRMOT}.
\end{abstract}


\section{Introduction} 
\label{sec:Introduction}

\IEEEPARstart{M}ulti-Object Tracking (MOT) is a fundamental technology for achieving advanced environmental understanding, playing an important role in fields such as autonomous driving \cite{chen2024delving, hou2026mmdrive, guan2025wavelet} and embodied intelligence \cite{wu2025cognitive, wang2025trackvla}. Recently, to enhance interactivity and semantic understanding, Referring Multi-Object Tracking (RMOT) \cite{wu2023referring} has emerged, transitioning the field from pure motion association to joint vision-language understanding. This task requires identifying and tracking specific objects guided by natural language, bridging the gap between semantics and dynamic environmental perception.

\begin{figure}[t]
\centering
    \includegraphics[width=1.0\linewidth]{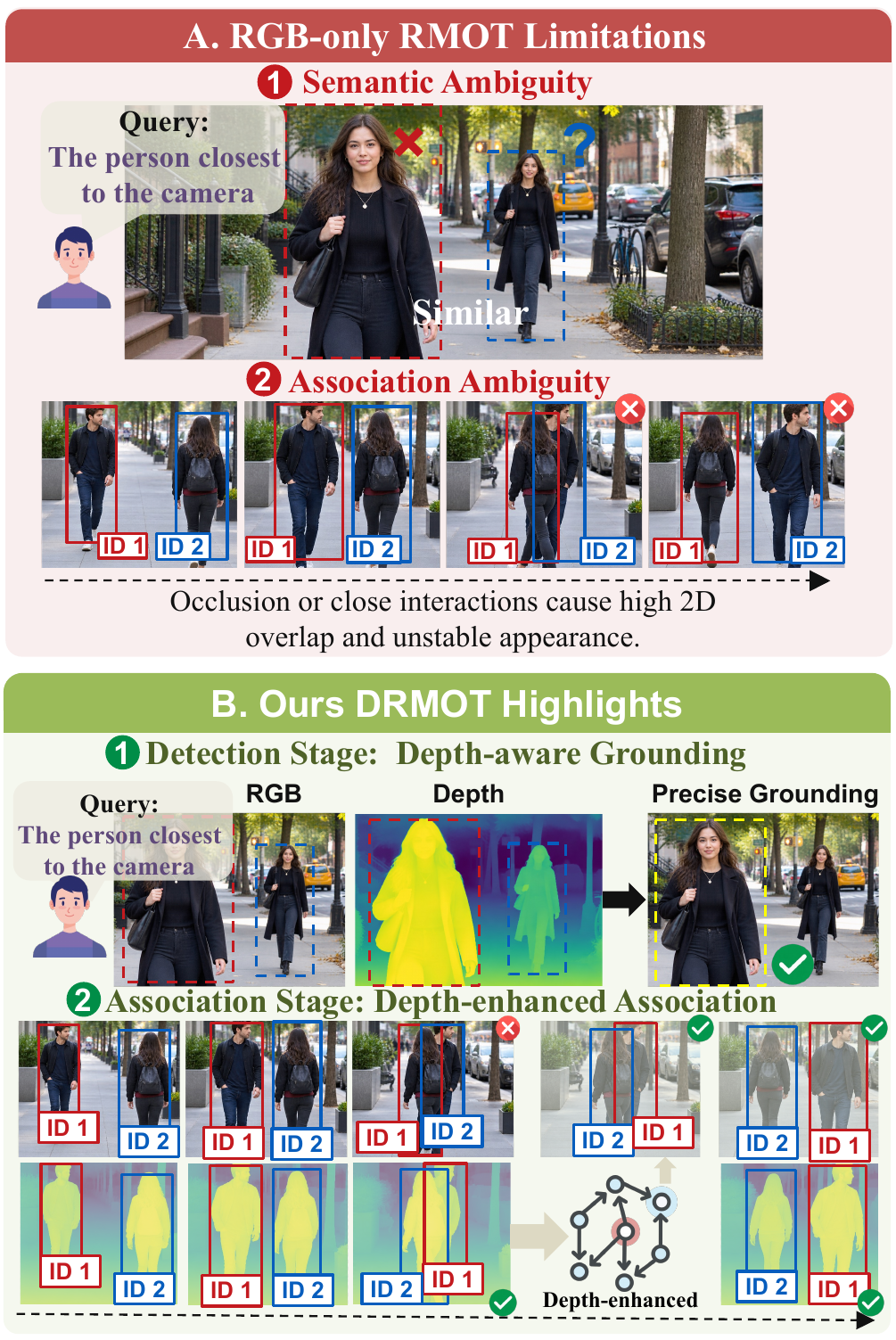}
    \caption{Comparison between RMOT and DRMOT tasks. (a) The RMOT model relying solely on language (L) description and RGB images is unable to correctly detect the referred target under depth-dependent spatial descriptions. (b) DRMOT leverages depth cues to resolve spatial ambiguity, thereby achieving accurate target detection and maintaining temporal identity consistency. }
    \label{fig: Comparison between RMOT and DRMOT.}
\end{figure}

As language-guided perception expands to embodied agents, interactive robots, and autonomous systems, RMOT is required not only to recognize objects but also to identify the object set specified by language in dynamic 3D environments and track these objects consistently over time. Existing RMOT formulations rely solely on 2D RGB observations, whereas referring expressions may specify metric distance, depth ordering, or occlusion relationships. \textbf{Because similar 2D projections can correspond to different 3D configurations, RGB evidence alone may not uniquely identify the referred targets.} This creates a fundamental mismatch between the task requirements and the available observations. As illustrated in Fig.~\ref{fig: Comparison between RMOT and DRMOT.}, this mismatch leads to two coupled failure modes. First, semantic ambiguity arises when visually similar candidates must be distinguished according to depth-dependent relations, resulting in incorrect target selection. Second, association ambiguity arises under occlusion or close-range interactions, where unstable appearance features and overlapping bounding boxes increase the risk of identity switches. Explicit depth addresses these failures in complementary ways: it supports language-conditioned geometric reasoning during detection and provides cross-frame geometric consistency during association. These distinct roles motivate the staged design of DRTrack, which combines depth-aware language grounding with depth-constrained trajectory association.

Based on the above analysis, we introduce a new task, termed RGB\textbf{D} \textbf{R}eferring \textbf{M}ulti-\textbf{O}bject \textbf{T}racking (\textbf{DRMOT}). Unlike conventional RMOT, DRMOT takes synchronized RGB images and depth maps as visual inputs. It also uses a language description that refers to one or more target objects. The goal is to localize all objects that satisfy the specified semantic and spatial constraints. The model must also maintain consistent identities throughout the video sequence. Therefore, it is required to produce accurate bounding boxes and temporally consistent track identities for all referred objects. The \textbf{core challenge} is to ground the language description in both visual appearance and 3D geometry. This enables accurate target selection and consistent identity preservation over time. As illustrated in Fig.~\ref{fig: Comparison between RMOT and DRMOT.}~(b), explicit depth helps resolve depth-dependent expressions that remain ambiguous in RGB projections, such as ``the person closest to the camera.'' It also provides additional geometric evidence for reliable association under occlusion.

To advance research on the DRMOT task, we construct \textbf{DRSet}, a novel RGB\textbf{D} \textbf{R}eferring multi-object tracking data\textbf{Set} designed to evaluate spatial-semantic detection and tracking in real-world 3D environments. Built upon the ARKitTrack \cite{zhao2023arkittrack} dataset, DRSet provides synchronized RGB images, pixel-wise depth maps, precise bounding boxes, and natural language descriptions. The dataset encompasses 186 diverse RGB-D video sequences (spanning outdoor and indoor environments, alongside normal and low-light conditions), systematically partitioned into 111 training and 75 testing sequences. To ensure semantic richness, DRSet contains 2,076 precise referring expressions with an average length of 5.86 words, of which 380 explicitly incorporate depth-dependent semantics (e.g., ``closer'', ``front'', ``behind''). Furthermore, the dataset introduces substantial real-world challenges through a broad spectrum of target categories (spanning humans, animals, vehicles, and objects), highly variable target densities (with up to 12 target identities per expression), dynamic action cues, and extended sequences reaching up to 1,479 frames. Generated via a rigorous multi-stage manual annotation and dual-annotator cross-review protocol, DRSet serves as a comprehensive and challenging validated resource for evaluating RGBD referring multi-object tracking models.

Furthermore, we propose \textbf{DRTrack}, a novel \textbf{D}epth-\textbf{R}eferring \textbf{Track}ing framework designed to achieve robust spatial-semantic detection and stable cross-frame association by tightly coupling MLLM-guided perception with geometric-constrained tracking. Specifically, the framework features a Depth-Promoted Language-Guided Detection stage, where a Multimodal Large Language Model (MLLM) \cite{bai2025qwen2} concurrently processes language description, RGB images, and depth maps to explicitly resolve depth-dependent semantics. To guarantee precise target localization and eliminate 2D spatial ambiguities, this MLLM is fine-tuned via Group Relative Policy Optimization (GRPO) \cite{shao2024deepseekmath} using specialized format and intersection-over-union (IoU) reward functions, enabling it to reliably output structured 2D bounding boxes. Subsequently, the Depth-Enhanced Association stage processes these spatial predictions by formulating an RGBD joint similarity metric that fuses traditional 2D IoU with a 3D depth consistency score, calculated via the exponential decay of absolute depth differences. As a result, DRTrack resolves identity confusion caused by severe occlusions or close 2D proximity, yielding accurate trajectories in complex scenes.

Extensive experiments on the DRSet dataset demonstrate that DRTrack achieves state-of-the-art (SOTA) performance. By leveraging synergistic L+RGB+D inputs and a parameter-efficient MLLM backbone, DRTrack attains a Higher Order Tracking Accuracy (HOTA) of 28.25\%, more than doubling the pure-RGB baseline. Ablation studies confirm that explicit depth integration and Geometric-Aware GRPO fine-tuning are crucial for accurately interpreting depth-dependent semantics and facilitating precise spatial-semantic detection. Furthermore, by optimally balancing 2D spatial context with 3D depth consistency, DRTrack exhibits remarkable association robustness against severe occlusions, effectively resolving complex spatial ambiguities and maintaining stable cross-frame trajectories in real-world 3D scenarios.

In summary, our main contributions are as follows:
\begin{itemize}
    \item We propose a novel task, \textbf{RGBD Referring Multi-Object Tracking (DRMOT)}, to overcome the inherent limitations of 2D-only RMOT methods in reasoning about complex spatial relationships. By jointly leveraging language, RGB, and depth modalities, this task establishes a new paradigm that requires models to ground language descriptions in both visual appearance and 3D geometry, effectively resolving semantic and association ambiguities caused by depth-dependent relations and occlusions.    
    \item We construct \textbf{DRSet}, a comprehensive dataset tailored for the DRMOT task. It encompasses 186 diverse RGB-D video sequences (partitioned into 111 training and 75 testing sequences) and provides 2,076 precise referring expressions. DRSet introduces substantial real-world challenges, including 380 expressions with explicit depth-dependent semantics, broad target categories, highly variable densities, and extended sequences.
    \item We propose \textbf{DRTrack}, an innovative MLLM-guided depth-referring tracking framework. It features an MLLM fine-tuned via Group Relative Policy Optimization (GRPO) for reliable bounding box prediction, alongside a Depth-Enhanced Association stage utilizing an RGBD joint similarity metric. Extensive experiments demonstrate that DRTrack achieves \textbf{state-of-the-art (SOTA)} performance (e.g., 28.25\% HOTA), more than doubling the pure-RGB baseline and highlighting the immense potential of synergizing explicit depth cues with MLLMs.
\end{itemize}

\section{Related Work}
\label{sec:Related Work}

\subsection{Referring Multi-Object Tracking}
Referring Multi-Object Tracking (RMOT) extends traditional Multi-Object Tracking (MOT) \cite{bewley2016simple, wojke2017simple, zhang2022bytetrack, cao2023observation, chen2024delving, li2024matching, wu2024temporal, zheng2025dfa, li2025ovtr, gao2025multiple, liang2025improved, huang2025dftrack, xue2025usvtrack, zhang2025lego} by incorporating natural language to selectively identify and track objects specified by a language description. Unlike generic MOT, which tracks all objects indiscriminately, RMOT requires joint vision-language understanding to localize targets based on language descriptions while maintaining consistent object identities over time. TransRMOT \cite{wu2023referring} pioneers this task by introducing the first RMOT benchmark and an end-to-end Transformer-based framework. Subsequent works explore different aspects of this multimodal challenge. For instance, some methods focus on bootstrapping temporal reasoning for complex expressions \cite{zhang2024bootstrapping,liang2025cognitive,chen2025cross}, while others emphasize practical deployment through plug-and-play and memory-efficient designs \cite{du2024ikun,tran2024mex}. Recently, large vision-language models have been explored for their zero-shot capabilities, enabling language-driven tracking without task-specific training \cite{chamiti2025refergpt,chen2025reamot}.
Despite these diverse and significant advances, existing RMOT-style methods remain largely limited to RGB-only observations. The absence of explicit 3D spatial modeling fundamentally restricts their ability to resolve complex depth-dependent ambiguities, underscoring the urgent necessity for introducing depth cues into the referring tracking paradigm.

\subsection{Referring Multi-Object Tracking Datasets}
Existing datasets for referring multi-object tracking (RMOT) have significantly advanced the field, yet they are predominantly constructed on RGB-only inputs. Refer-KITTI \cite{wu2023referring} pioneered this line of research by providing the first dedicated benchmark for RMOT in autonomous driving scenarios, utilizing language to guide the tracking of multiple vehicles and pedestrians. Refer-Dance \cite{du2024ikun} emphasizes highly dynamic human movements, while LaMOT \cite{li2025lamot} places a greater focus on temporal complexity and long-term trajectory associations. But a critical limitation persists: the complete reliance on 2D visual observations. The absence of explicit depth information restricts these datasets from effectively supporting or evaluating depth-dependent referring expressions. As a result, existing benchmarks are insufficient for rigorously assessing a model's true spatial-semantic detection capabilities under severe geometric ambiguity. This fundamental gap underscores the urgent need for multimodal benchmarks that seamlessly fuse visual, linguistic, and 3D geometric information, thereby motivating our construction of the DRSet dataset.

\subsection{Depth-based Perception}
Depth information provides essential geometric cues that bridge the gap between 2D visual observations and 3D spatial understanding, offering explicit structural priors that are highly resilient to illumination changes and appearance variations. In the realm of visual tracking, depth has been widely adopted to resolve scale ambiguities, stabilize object localization, and enforce spatial consistency, proving especially crucial under severe occlusions and drastic viewpoint changes \cite{pang2023standing, yan2024monocd}. CDTB \cite{lukezic2019cdtb} pioneered the comprehensive evaluation of RGB-D trackers by introducing diverse indoor and outdoor scenarios. Subsequently, DepthTrack \cite{yan2021depthtrack} significantly expanded the scale and scene complexity of available data, providing a robust benchmark for long-term depth-aware tracking. More recently, ARKitTrack \cite{zhao2023arkittrack} leveraged consumer-grade LiDAR to capture high-fidelity RGB-D sequences, further driving the development of trackers that tightly couple 3D spatial geometry with 2D visual appearance. Collectively, these studies and benchmarks demonstrate that depth serves as a versatile and robust geometric prior in complex dynamic scenes, which directly motivates our integration of explicit depth perception into the referring multi-object tracking paradigm.

\subsection{Multimodal Large Language Models} 

The emergence of Multimodal Large Language Models (MLLMs) represents a profound paradigm shift toward general-purpose multimodal comprehension \cite{fei2024multimodal}.  Pioneering this advancement, LLaVA-1.5 \cite{liu2024visual} establishes a robust foundation for visual instruction tuning, enabling LLMs to perceive and converse about image content. Building upon this, LLaVA-NEXT \cite{liu2024llavanext} further enhances visual resolution and dynamic context reasoning, allowing for more nuanced image interpretation. Concurrently, DeepSeek-VL \cite{lu2024deepseek} introduces a highly efficient architecture that excels in real-world visual-language alignment and diverse multimodal dialogue. Cutting-edge models such as Qwen2.5-VL \cite{bai2025qwen2} and InternVL3 \cite{zhu2025internvl3} push the boundaries of instance-level object detection and spatial reasoning.  Consequently, they provide a technical foundation for dense prediction tasks, particularly referring multi-object tracking, where fine-grained spatial-semantic alignment and robust cross-modal reasoning are paramount.

\begin{figure*}[t]
    \centering
    \includegraphics[width=1.0\linewidth]{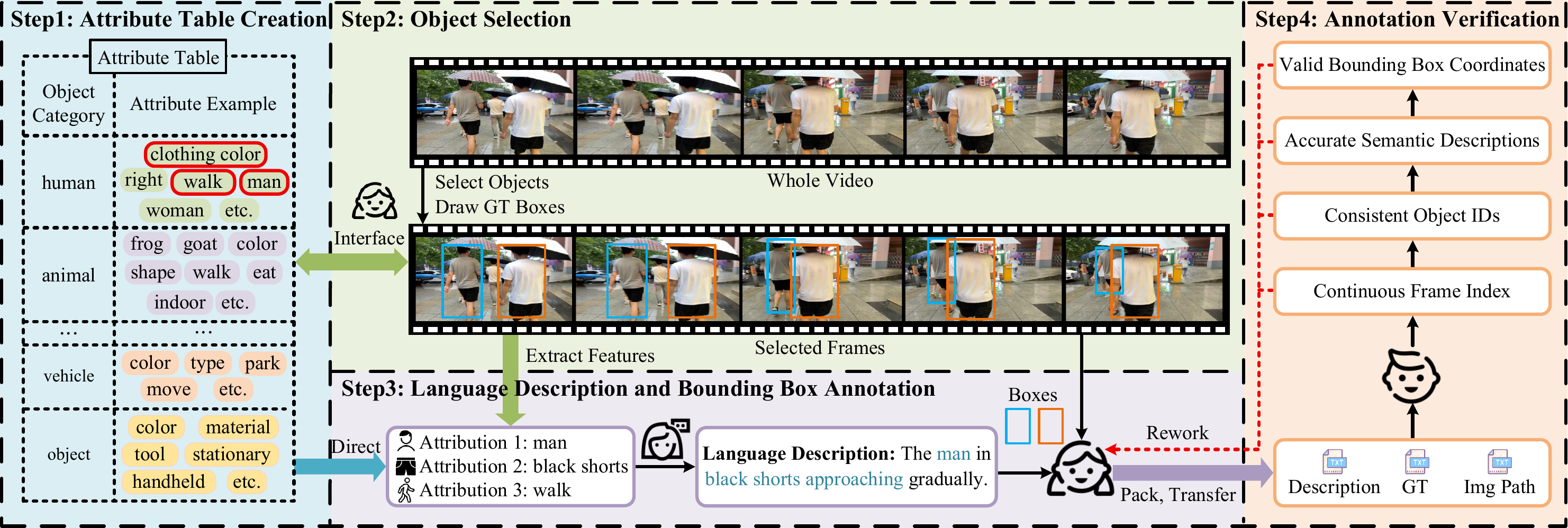}
    \caption{Annotation Process. The annotation process includes four steps: (1) Attribute Table Creation: we build an attribute table that categorizes object descriptions into static attributes and dynamic behaviors; (2) Object Selection: we review the entire video and select representative targets based on their attributes and behaviors; (3) Language Description Annotation: we draw bounding boxes frame by frame and compose language descriptions according to the predefined attributes; and (4) Annotation Verification: we perform a cross-review by dual-annotator to ensure valid bounding box coordinates, consistent object IDs, and accurate language descriptions. Finally, all verified annotations are packed to generate the DRSet dataset.}
    \label{fig:annotation_process.}
\end{figure*}

\section{Dataset}
\label{sec:Dataset}

To advance RGBD Referring Multi-Object Tracking, we expand ARKitTrack into a multimodal benchmark that couples language descriptions with synchronized RGB images, depth maps, and frame-level target boxes. A comparison with existing tracking datasets is provided in Table \ref{tab:dataset_comparison.}.

\subsection{Dataset Collection}
The core components of our DRSet dataset comprise the language, RGB, and depth modalities. Specifically, the RGB modality provides rich visual appearance cues, while the depth modality (D) offers pixel-wise distance measurements from object surfaces to the camera’s optical center. In addition, the language modality supplies natural-language descriptions that semantically characterize each sample, facilitating cross-modal understanding. To construct such a dataset, we first survey and compare existing RGB-D object tracking datasets, including CDTB \cite{lukezic2019cdtb}, DepthTrack \cite{yan2021depthtrack}, and ARKitTrack \cite{zhao2023arkittrack}. After comparing these datasets in terms of sequence quantity, scene complexity, and depth availability, we adopt ARKitTrack as the foundational dataset for building DRSet. Building upon its high-quality RGB-D data, we further extend it with language descriptions and multi-object bounding box annotations, thereby constructing a multimodal RGB-D dataset for referring multi-object tracking, termed DRSet.

\subsection{Dataset Annotation Process}
To ensure high-quality language descriptions and accurate bounding boxes, we manually annotate the targets in each sequence. The annotation process is illustrated in Fig. \ref{fig:annotation_process.}, with specific steps outlined below:

\subsubsection{Attribute Table Creation}
We divide the content of the language descriptions into two main categories: static attribute descriptions and dynamic behavior descriptions, to systematically annotate the targets. Each category is further subdivided into several attributes, as shown in Table \ref{tab:attribute table}.

\begin{table}[t]
\caption{Comparison of existing tracking datasets and the expanded DRSet dataset.}
\label{tab:dataset_comparison.}
\resizebox{1.0 \linewidth}{!}{
    \centering
    \begin{tabular}{lcccccc}
        \toprule
        Tracking Dataset & Language & RGB & Depth & Sequences & Expressions & Avg.Words \\
        \midrule
        Refer-KITTI \cite{wu2023referring} & \textcolor[HTML]{00B050}{\ding{51}} & \textcolor[HTML]{00B050}{\ding{51}} & \textcolor{red}{\ding{55}} & 18 & 818 & 49 \\
        Refer-Dance \cite{du2024ikun} & \textcolor[HTML]{00B050}{\ding{51}} & \textcolor[HTML]{00B050}{\ding{51}} & \textcolor{red}{\ding{55}} & 65 & 1.9K & 25 \\
        LaMOT \cite{li2025lamot} & \textcolor[HTML]{00B050}{\ding{51}} & \textcolor[HTML]{00B050}{\ding{51}} & \textcolor{red}{\ding{55}} & 62 & 145 & 9 \\
        \midrule
        CDTB \cite{lukezic2019cdtb} & \textcolor{red}{\ding{55}} & \textcolor[HTML]{00B050}{\ding{51}} & \textcolor[HTML]{00B050}{\ding{51}} & 80 & -- & -- \\ 
        DepthTrack \cite{yan2021depthtrack} & \textcolor{red}{\ding{55}} & \textcolor[HTML]{00B050}{\ding{51}} & \textcolor[HTML]{00B050}{\ding{51}} & 200 & -- & -- \\
        ARKitTrack \cite{zhao2023arkittrack} & \textcolor{red}{\ding{55}} & \textcolor[HTML]{00B050}{\ding{51}} & \textcolor[HTML]{00B050}{\ding{51}} & 300 & -- & -- \\
        \midrule
        \rowcolor[HTML]{e2f0ff} \textbf{DRSet (Ours)} & \textbf{\textcolor[HTML]{00B050}{\ding{51}}} & \textbf{\textcolor[HTML]{00B050}{\ding{51}}} & \textbf{\textcolor[HTML]{00B050}{\ding{51}}} & 186 & 2,076 & 5.86 \\
        \bottomrule
    \end{tabular}
}
\end{table}

\begin{table}[t]
\caption{Examples of attribute dimensions and values for different subject types.}
\label{tab:attribute table}
\resizebox{1.0\linewidth}{!}{
    \centering
    \begin{tabular}{cll}
        \toprule
        Subject Type & Attribute Dimension & Example Attribute Values \\
        \midrule
        \multirow{6}{*}{Human} 
            & Clothing & Clothing color, clothing type, accessories \\
            & Hand-held objects & Backpack, basketball, umbrella, etc. \\
            & Gender / Age & Male, female, child \\
            & Actions & Waving, walking, eating, standing, etc. \\
            & Scene interaction & Leaving, approaching \\
            & Spatial position & Right side, front, back \\
        \midrule
        \multirow{4}{*}{Animal} 
            & Species & Anteater, dragonfly, frog, goat, etc. \\
            & Appearance features & Color, body shape, local features \\
            & Actions & Standing, walking, eating, lying down, etc. \\
            & Habitat / Scene & Indoor, grassland, water area, etc. \\
        \midrule
        \multirow{2}{*}{Vehicle} 
            & Appearance features & Color, type of vehicle \\
            & Motion state & Parking, moving \\
        \midrule
        \multirow{3}{*}{Object} 
            & Category & Electronic devices, tools, containers, etc. \\
            & Appearance features & Color, material, brand \\
            & Status & Static, rolling, hand-held \\
        \bottomrule
    \end{tabular}
}
\end{table}

\subsubsection{Object Selection} 
We review each video to identify potential tracking targets based on their static attributes and dynamic behaviors. We prioritize targets that satisfy the tracking requirements and possess depth information, ensuring both diversity and representativeness of the dataset.

\subsubsection{Language Description and Bounding Box Annotation}
We first conduct detailed labeling of targets sharing the same attributes based on a predefined attribute table, and then compose language descriptions for these targets. Finally, we use an annotation tool to draw accurate bounding boxes frame-by-frame for the selected targets. During this process, annotations are recorded in a structured format, where each entry contains the frame number (fn), identity (id), and bounding box coordinates $[x_1, y_1, x_2, y_2]$, explicitly describing the position and size of each target. Meanwhile, we record auxiliary metadata such as image paths and ground-truth language descriptions.

\subsubsection{Annotation Verification}
To ensure annotation accuracy, we employ a dual-annotator cross-review protocol. Following the completion of the initial annotation, a second independent annotator reviews the results to eliminate potential mislabels and inconsistencies between the targets' bounding boxes and the language descriptions. The annotation content that passes the review will be integrated and packaged, while the content that fails the review will be returned for revision. Ultimately, all verified annotation data is integrated to generate the dataset.

\subsection{Dataset Split}

We split the DRSet dataset into disjoint training and testing subsets at the video level to avoid scene overlap. In total, DRSet contains 1,324 training expressions and 752 test expressions, corresponding to 63.78\% and 36.22\% of all expressions. The training and test partitions cover 111 and 75 RGB-D source sequences, respectively.

\begin{figure*}[t]
    \centering
    \includegraphics[width=1.0\linewidth]{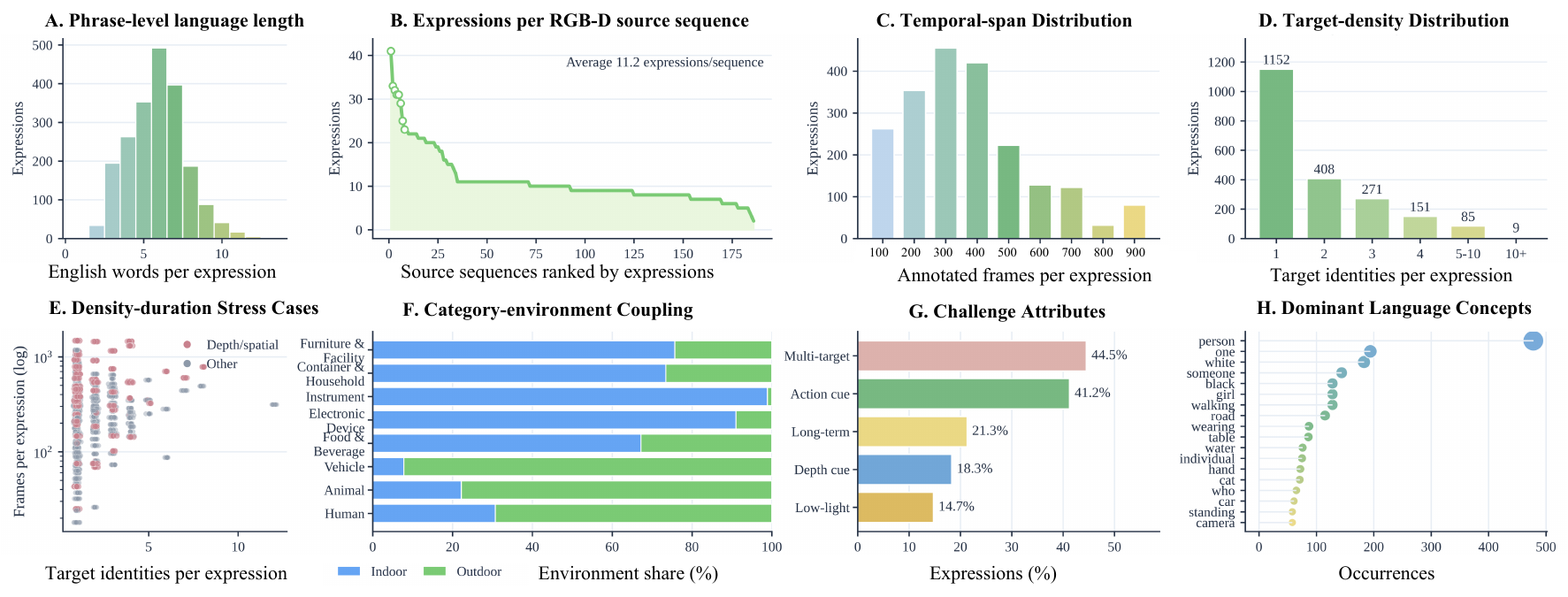}
    \caption{Overview of the DRSet dataset statistics. (A) Distribution of English words per referring expression. (B) Expressions per RGB-D source sequence. (C) Distribution of annotated temporal spans. (D) Distribution of target identities per expression. (E) Joint distribution of target density and temporal span, with depth and spatial expressions highlighted. (F) Environment composition within major semantic categories. (G) Proportions of key challenge attributes. (H) Dominant language concepts in the referring expressions.}
    \label{fig:dataset}
\end{figure*}

\subsection{Dataset Statistics}

The  DRSet contains 2,076 referring expressions over 186 RGB-D source sequences. The annotation records provide 711.3K expression-level annotated frames, 966.2K expression-level bounding-box annotations, and 4,017 expression-level target tracks. The underlying RGB-D source pool contains 244.6K paired RGB/depth frames. The dataset covers outdoor and indoor environments, which account for 61.75\% and 38.25\% of the expressions, respectively. The current metadata-based grouping assigns 85.26\% of the expressions to normal lighting and 14.74\% to low-light conditions. It assigns 81.50\% to Dynamic and 18.50\% to Static-camera/object-biased motion states.  We conduct a comprehensive analysis of the DRSet dataset from multiple perspectives, as follows:

As shown in Fig.~\ref{fig:dataset}-(A), the referring expressions are concise, with an average length of 5.86 words. Although compact, these expressions cover target categories, appearance attributes, actions, and spatial relations. Fig.~\ref{fig:dataset}-(H) further shows that the vocabulary includes person-centered concepts, visual attributes, motion descriptions, and scene-related terms. This language distribution supports practical referring scenarios where models must ground short expressions in complex RGB-D videos. Fig.~\ref{fig:dataset}-(B) characterizes the number of annotated referring expressions per RGB-D sequence. Each sequence contains 11.2 referring expressions on average. Specifically, a single RGB-D sequence may contain multiple valid targets, spatial relations, and language conditions. This design increases the diversity of referring cases within each scene.

Fig.~\ref{fig:dataset}-(C) demonstrates that DRSet covers a broad range of temporal spans. Each expression contains 342.62 annotated frames on average (median: 288; maximum: 1,479). While many expressions fall within the 100--500 frame range, the dataset also includes long-term cases exceeding 1,000 frames. This temporal coverage enables evaluation under both short-term localization and long-term association settings.

The target-density statistics in Fig.~\ref{fig:dataset}-(D) highlight the multi-object nature of DRSet. Single-target expressions account for 55.49\% of the benchmark, whereas 924 expressions (44.51\%) involve multiple target identities, with up to 12 identities per expression. The joint distribution in Fig.~\ref{fig:dataset}-(E) reveals that target density and temporal span co-vary. The 856 expressions with action or dynamic related terms, corresponding to 41.23\%, and 380 expressions with depth-related or spatial-related terms, corresponding to 18.30\%. These proportions indicate the presence of relevant language cues under the current matching rules rather than independent human challenge labels. Expressions containing terms such as ``front'', ``behind'', ``closer'', ``near'' or ``distance'' can remain ambiguous in 2D projections. The depth modality provides additional geometric evidence for resolving these cases.

Fig.~\ref{fig:dataset}-(F) presents the environment composition across major semantic categories. Outdoor and indoor environments constitute 61.75\% and 38.25\% of all expressions, respectively. The category-level distribution spans humans, animals, vehicles, facilities, household objects, devices, instruments, and food-related items across diverse scene contexts. This breadth strengthens DRSet's ability to evaluate generalization across semantic categories and physical environments.  Fig.~\ref{fig:dataset}-(G) summarizes representative challenge attributes in DRSet: multi-target expressions (44.5\%), action-related cues (41.2\%), long-term cases (21.3\%), depth/spatial cues (18.3\%), and low-light conditions (14.7\%). These attributes encompass key challenges in RGB-D referring multi-object tracking, including target interaction, motion variation, spatial ambiguity, long-term temporal continuity, and illumination changes. Collectively, DRSet provides a comprehensive and effective benchmark for evaluating language-guided RGB-D multi-object tracking in realistic and diverse scenarios.

These statistics reveal that DRSet is a demanding benchmark. Multiple referred identities frequently coexist within long sequences, the language often encodes depth-dependent relations rather than appearance alone, and a non-negligible portion of the data is captured under poor illumination. These factors overlap within the same sequence, so a tracker must simultaneously handle semantic grounding, geometric reasoning, and long-term identity preservation.

\begin{figure*}[t]
    \centering
    \includegraphics[width=1.0\linewidth]{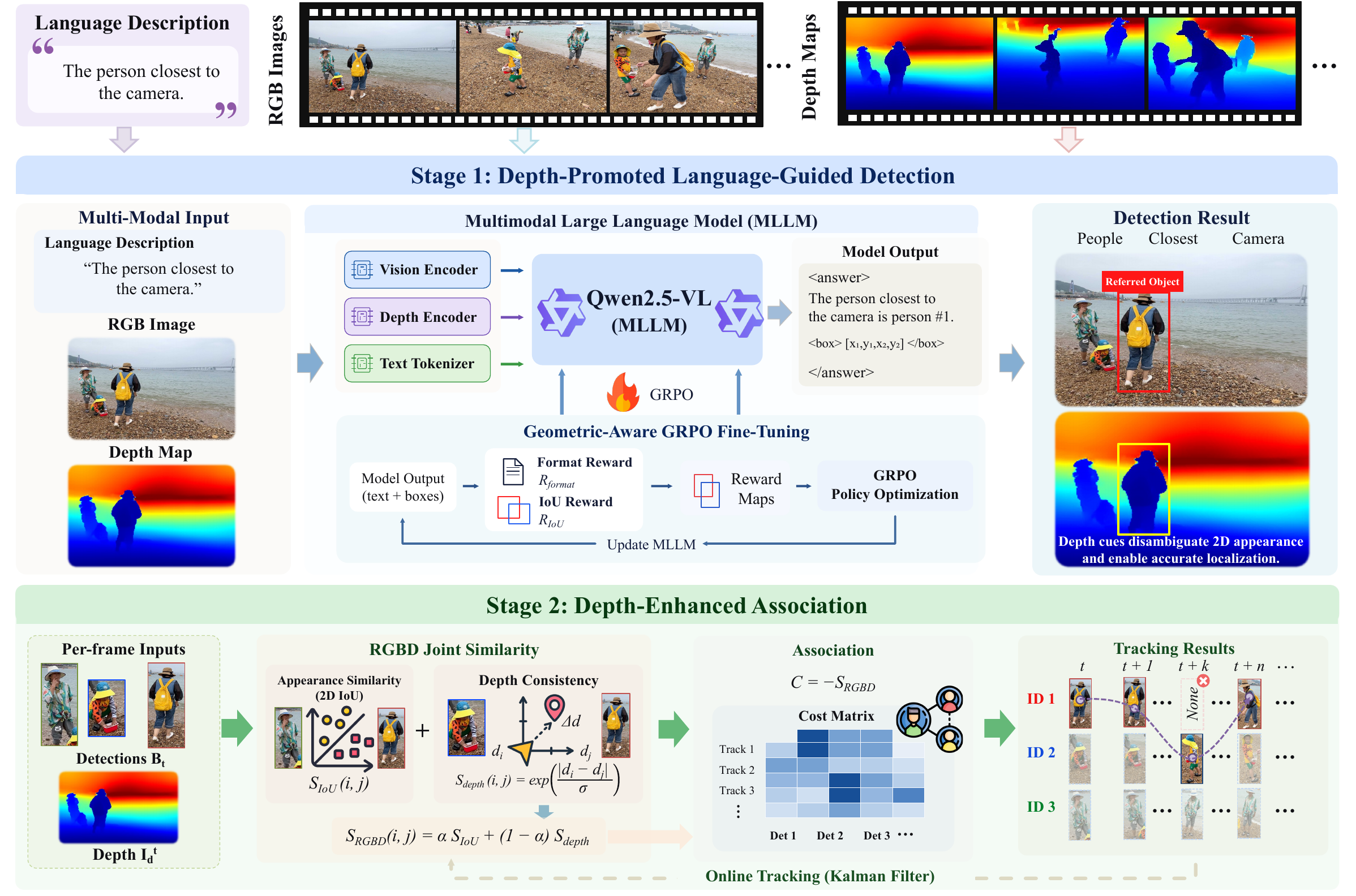}
    \caption{Overall pipeline of DRTrack. DRTrack consists of two stages. In Stage 1, the Depth-Promoted Language-Guided Detection stage leverages an MLLM to jointly process language descriptions, RGB images, and depth maps, and is fine-tuned via Geometric-Aware GRPO with format and IoU rewards to produce accurate depth-aware 2D bounding boxes of referred objects. In Stage 2, the Depth-Enhanced Association stage integrates depth information to compute the RGBD joint similarity as the association cost, thereby yielding ID-consistent trajectories of the referred targets.}
    \label{fig:overall pipeline of DRTrack.}
\end{figure*}

\section{Methodology}
\label{sec:methodology}

The overall architecture of the proposed DRTrack framework is designed to address the unique challenges in RGBD Referring Multi-Object Tracking (DRMOT). As illustrated in Fig. \ref{fig:overall pipeline of DRTrack.}, the framework tightly couples a Multimodal Large Language Model (MLLM) with a robust geometric tracking pipeline. DRTrack comprises two stages: (1) \textbf{Depth-Promoted Language-Guided Detection}, which enhances the MLLM's ability to localize specific targets by leveraging depth-specific semantics in language descriptions, and (2) \textbf{Depth-Enhanced Association}, which maintains target identities across frames using a combination of multimodal spatial-geometric cues and an online trajectory smoothing mechanism.

\subsection{Depth-Promoted Language-Guided Detection}

This stage uses the multimodal understanding capabilities of an MLLM to align the language descriptions with potential referred objects in the RGB-D scene. The core objective is to leverage the metric geometry of the depth map to enhance the MLLM's comprehension of depth-dependent spatial relationships (e.g., ``the closest car'', ``the cat that is closer to the tree'') to achieve accurate 2D bounding box localization.

\subsubsection{Inputs and Outputs}

The detection module concurrently processes three distinct modalities to guarantee precise semantic-grounded target localization. The unified multimodal input is formulated as:
\begin{equation}
\mathcal{X} = \{L, I_{\text{RGB}}, \phi(I_{\text{D}})\}
\end{equation}
where $L$ denotes the language description specifying target categories and their depth-related attributes, $I_{\text{RGB}} \in \mathbb{R}^{H \times W \times 3}$ represents the RGB image providing appearance cues, $I_{\text{D}} \in \mathbb{R}^{H \times W \times 1}$ is the depth map providing explicit metric distance information, and $\phi(\cdot)$ is a transformation function that converts the single-channel depth map into a 3-channel pseudo-RGB representation to match the visual backbone's interface.

By concurrently feeding these modalities, the cross-modal features are fused to generate a structured text sequence:
\begin{equation}
T = \mathcal{F}_{\text{MLLM}}(\mathcal{X})
\end{equation}
where $\mathcal{F}_{\text{MLLM}}$ denotes the forward inference process of the Multimodal Large Language Model, and $T$ is the generated text sequence output.

Finally, a predefined parsing operation extracts the precise 2D bounding box coordinates of the referred objects from the generated text:
\begin{equation}
\mathbf{B} = \psi(T) = \left\{ [x_1^{(i)}, y_1^{(i)}, x_2^{(i)}, y_2^{(i)}] \right\}_{i=1}^{N}
\end{equation}
where $\psi(\cdot)$ denotes the parsing extraction function, $\mathbf{B}$ is the set of extracted bounding boxes, $[x_1^{(i)}, y_1^{(i)}, x_2^{(i)}, y_2^{(i)}]$ are the top-left and bottom-right coordinates of the $i$-th target, and $N$ denotes the total number of detected referred targets.

\subsubsection{Geometric-Aware GRPO Fine-Tuning}

While generic Multimodal Large Language Models (MLLMs) excel at visual-textual alignment, they are not inherently optimized for precise geometric localization. Conventional Supervised Fine-Tuning (SFT) primarily optimizes token-level cross-entropy, which inherently struggles to capture continuous, non-differentiable spatial metrics (e.g., Intersection-over-Union (IoU)) and lacks the exploratory capacity required for complex spatial reasoning. To overcome these limitations, we employ the Group Relative Policy Optimization (GRPO) algorithm \cite{shao2024deepseekmath} to fine-tune the MLLM. GRPO shifts the learning paradigm from passive token imitation to active spatial exploration, directly optimizing the model's policy using geometric feedback without requiring a computationally expensive Critic network.

The fine-tuning process is driven by a composite reward function that jointly enforces structure and accuracy:
\begin{equation}
R_{\text{total}} = R_{\text{format}} + R_{\text{IoU}}
\end{equation}
where $R_{\text{total}}$ is the total reward for the generated response, $R_{\text{format}}$ is the format reward, and $R_{\text{IoU}}$ is the IoU reward.

The \textbf{Format Reward} $R_{\text{format}}$ guarantees that the model produces strictly parsable and structured text schemas (i.e., {\ttfamily \textless think\textgreater...\textless /think\textgreater \textless answer\textgreater{JSON}\textless /answer\textgreater}), which is a prerequisite for robust automated coordinate extraction. Formally,
\begin{equation}
R_{\text{format}} =
\begin{cases}
1, & \text{if format is correct} \\
0, & \text{if format is incorrect}
\end{cases}
\end{equation}

The \textbf{IoU Reward} $R_{\text{IoU}}$ provides a direct, non-differentiable geometric optimization signal to fine-tune the spatial regression capabilities:
\begin{equation}
R_{\text{IoU}} = \sum_{(\text{i, j}) \in \text{Matches}} \text{IoU}(\text{Box}_{\text{pred}, \text{i}}, \text{Box}_{\text{gt}, \text{j}})
\end{equation}
where $\text{Box}_{\text{pred}, \text{i}}$ and $\text{Box}_{\text{gt}, \text{j}}$ represent the $i$-th predicted and $j$-th ground-truth boxes, respectively. The optimal one-to-one correspondences between the $M$ predicted boxes and $N$ ground-truth boxes are established by executing the Hungarian algorithm over the negative pairwise IoU matrix. By optimizing relative group advantages under $R_{\text{total}}$, GRPO forces the MLLM to autonomously co-activate its inner reasoning steps within the {\ttfamily \textless think\textgreater} tags and generate pixel-perfect coordinate boundaries within the {\ttfamily \textless answer\textgreater} tags.

\subsection{Depth-Enhanced Association}
\label{subsection:depth_enhanced_association}

Following target localization, the tracking stage employs a streamlined, depth-augmented data association mechanism to maintain stable target identities across frames while minimizing computational overhead.

\subsubsection{Interval Inference and Online Trajectory Smoothing}

To balance computational efficiency and tracking robustness, our framework employs an interval-based inference strategy. The MLLM performs language-guided detection strictly every $K$ frames to provide authoritative target localizations. 

The trajectory state at frame $t$ is modeled by a 7-dimensional vector $\mathbf{x}_t = [x, y, s, r, \dot{x}, \dot{y}, \dot{s}]^\top$. For the intermediate frames where $t \pmod K \neq 0$ and direct MLLM observations are absent, the framework propagates the trajectory state using a Constant Velocity Kalman Filter model:
\begin{equation}
\mathbf{x}_{t|t-1} = \mathbf{F} \mathbf{x}_{t-1|t-1}
\end{equation}
\begin{equation}
\mathbf{P}_{t|t-1} = \mathbf{F} \mathbf{P}_{t-1|t-1} \mathbf{F}^\top + \mathbf{Q}
\end{equation}
where $\mathbf{x}_{t|t-1}$ is the predicted state vector, $(x, y)$ is the 2D bounding box center, $s$ denotes the scale (area), $r$ signifies the aspect ratio, and $(\dot{x}, \dot{y}, \dot{s})$ are their respective first-order time derivatives. $\mathbf{F} \in \mathbb{R}^{7 \times 7}$ is the state transition matrix, $\mathbf{P}$ is the state covariance matrix, and $\mathbf{Q}$ denotes the process noise covariance matrix. During these unobserved intervals, the posterior state defaults to the predicted state ($\mathbf{x}_{t|t} = \mathbf{x}_{t|t-1}$).

Importantly, to eliminate the cumulative estimation drift during these unobserved gaps, we introduce an online trajectory smoothing mechanism via virtual observations. Suppose a trajectory loses authoritative MLLM observations at frame $t_1$ and successfully re-associates with a new valid MLLM detection at frame $t_2$. To ensure geometric consistency during interpolation, the scale and aspect ratio are first mapped back to the physical width and height via:
\begin{equation}
w_i = \sqrt{s_i \cdot r_i}, \quad h_i = \sqrt{s_i / r_i}, \quad \text{for } i \in \{1, 2\}
\end{equation}
where $w_i$ and $h_i$ denote the width and height of the bounding box at frame $t_i$, derived from the scale $s_i$ and aspect ratio $r_i$.

Assuming a constant speed hypothesis across the unobserved interval $\Delta t = t_2 - t_1$, the linear velocity gradients for the bounding box parameters are formulated as:
\begin{equation}
\begin{aligned}
\delta_x = \frac{x_2 - x_1}{\Delta t}, \quad \delta_y = \frac{y_2 - y_1}{\Delta t} \\
\delta_w = \frac{w_2 - w_1}{\Delta t}, \quad \delta_h = \frac{h_2 - h_1}{\Delta t}
\end{aligned}
\end{equation}
where $\delta_x, \delta_y, \delta_w$, and $\delta_h$ denote the frame-by-frame rate of change for the center coordinates, width, height, respectively.

For any intermediate missing frame $\tau$ where $t_1 < \tau < t_2$, a virtual bounding box representation is constructed by uniform linear interpolation:
\begin{equation}
\begin{aligned}
x_\tau &= x_1 + (\tau - t_1) \delta_x, \quad y_\tau = y_1 + (\tau - t_1) \delta_y \\
w_\tau &= w_1 + (\tau - t_1) \delta_w, \quad h_\tau = h_1 + (\tau - t_1) \delta_h
\end{aligned}
\end{equation}
where $x_\tau, y_\tau, w_\tau$, and $h_\tau$ are the interpolated geometric properties of the virtual bounding box at frame $\tau$.

Thus, the virtual measurement vector is reconstructed for the missing frame by re-calculating the scale and aspect ratio:
\begin{equation}
s_\tau = w_\tau \cdot h_\tau, \quad r_\tau = \frac{w_\tau}{h_\tau}
\end{equation}
where $s_\tau$ and $r_\tau$ are the reconstructed scale and aspect ratio, forming the virtual measurement $\mathbf{z}_\tau = [x_\tau, y_\tau, s_\tau, r_\tau]^\top$.

Once the sequence of virtual measurements $\{\mathbf{z}_\tau\}_{\tau=t_1+1}^{t_2-1}$ is generated, the tracker unfreezes its historical state at $t_1$ and retroactively executes sequential Kalman predict-update cycles to refresh its internal parameters:
\begin{equation}
\mathbf{K}_\tau = \mathbf{P}_{\tau|\tau-1} \mathbf{H}^\top (\mathbf{H} \mathbf{P}_{\tau|\tau-1} \mathbf{H}^\top + \mathbf{R})^{-1}
\end{equation}
\begin{equation}
\mathbf{x}_{\tau|\tau} = \mathbf{x}_{\tau|\tau-1} + \mathbf{K}_\tau (\mathbf{z}_\tau - \mathbf{H} \mathbf{x}_{\tau|\tau-1})
\end{equation}
\begin{equation}
\mathbf{P}_{\tau|\tau} = (\mathbf{I} - \mathbf{K}_\tau \mathbf{H}) \mathbf{P}_{\tau|\tau-1}
\end{equation}
where $\mathbf{K}_\tau$ is the Kalman gain at frame $\tau$, $\mathbf{H} \in \mathbb{R}^{4 \times 7}$ represents the measurement matrix mapping the state space to the measurement space, and $\mathbf{R}$ is the measurement noise covariance. This retroactive online smoothing process effectively calibrates the intermediate state deviations.

\subsubsection{RGBD Joint Similarity Metric}

The core of our data association lies in the formulation of the RGBD Joint Similarity ($S_{\text{RGBD}}$), which elegantly integrates 2D spatial overlaps with 3D geometric consistency.

For each extracted 2D detection and predicted trajectory bounding box, the 3D geometric consistency is quantified via a depth similarity term $S_{\text{D}}$, defined as an exponential decay function of their absolute depth discrepancy:
\begin{equation}
S_{\text{D}} = \exp\left(-\frac{|D_{det} - D_{trk}|}{\sigma}\right)
\end{equation}
where $D_{det}$ and $D_{trk}$ are the mean depth values sampled from the valid pixels within the detection and tracking bounding boxes on the current frame's depth map, respectively, and $\sigma$ represents the depth scaling factor that regulates the sensitivity to depth differences.

The RGBD joint similarity metric is computed as a weighted sum of the 2D spatial Intersection-over-Union (IoU) and the 3D depth similarity $S_{\text{D}}$:
\begin{equation}
\label{eq:RGBD joint similarity.}
S_{\text{RGBD}} = \alpha \cdot \text{IoU} + (1 - \alpha) \cdot S_{\text{D}}
\end{equation}
where $\alpha \in [0, 1]$ denotes the similarity fusion weight, balancing the visual spatial overlap and metric geometric distance.

\subsubsection{Association Cost Function}

Unlike some tracking algorithms that rely heavily on complex motion priors, our framework utilizes a streamlined global assignment approach driven by spatial and geometric alignment. To formulate the matching task as a standard minimization problem, we construct the cost matrix $C$ by directly negating the RGBD joint similarity:
\begin{equation}
\label{eq:cost matrix}
C = - S_{\text{RGBD}}
\end{equation}
where $C$ represents the pairwise assignment costs between current detections and active tracks.

With the cost matrix established, we employ the Hungarian algorithm \cite{kuhn1955hungarian} to compute the optimal bipartite matching that minimizes the total global cost. This formulation effectively maximizes the joint similarity while avoiding the brittleness of heuristic motion vectors. Subsequent trajectory management, which includes track initialization, recovery, and termination, strictly follows a standard tracking life cycle.


\section{Experiments}
\label{sec:Experiments}

\subsection{Experimental Setup}

DRTrack is implemented with Qwen2.5-VL-3B-Instruct \cite{bai2025qwen2} as the default MLLM backbone. Unless otherwise specified, the reported configuration takes synchronized language, RGB, and depth inputs, applies the RGBD joint similarity in Equation (\ref{eq:RGBD joint similarity.}), and uses the MLLM after Geometric-Aware GRPO fine-tuning. The depth scaling factor is set to $\sigma=20$ and the similarity fusion weight to $\alpha=0.3$, which places a larger weight on depth consistency while retaining 2D spatial overlap as a complementary cue. The MLLM is invoked every 10 frames by default, and the Kalman filter with online trajectory smoothing propagates the tracks in the intermediate frames.

For parameter-efficient adaptation, only the LoRA parameters are fine-tuned and the vision encoder is kept frozen. The LoRA rank, scaling factor, and dropout rate are set to $64$, $128$, and $0.05$, respectively, and the learning rate is $1 \times 10^{-5}$. The GRPO stage uses $10\%$ of the DRSet training data and samples four responses for each training instance. The policy is optimized with the composite reward in Section \ref{sec:methodology}, and the coefficient of the KL regularization term is set to $0.04$. All experiments are conducted on a single NVIDIA A6000 GPU with a batch size of 4 and 2 gradient accumulation steps.

The prompts are specified according to the input setting. All zero-shot MLLM baselines,  use a generic detection prompt that asks the model to detect all instances matching the language description and to return their bounding boxes in the $[x1, y1, x2, y2]$ format. Since the same prompt is used for both L+RGB and L+RGB+D inputs, the comparison between these two settings reflects only the effect of the additional depth image. Each GRPO-adapted model is evaluated with the same prompt used during its training. For L+RGB+D inputs, this prompt states that aligned RGB and depth images from the same timestamp are provided and asks the model to infer all regions referred to by the instruction. 

All compared RMOT trackers are evaluated with their publicly released implementations and weights. The RGB-D multi-object trackers are combined with iKUN \cite{du2024ikun} to select the referred tracks, with the iKUN threshold set to $0.1$ for the YOLOX-based variants and to $0$ for the RGB-D variants. The MLLM-based baselines use SORT \cite{bewley2016simple} for association and share the 10-frame inference interval of DRTrack, whereas the remaining trackers process every frame.

\begin{table*}[t]
\centering
\caption{Performance of different methods on the test set of the DRSet dataset. ↑ indicates that higher score is better. The best results are marked in \textbf{bold}.}
\label{tab:main_results}
\resizebox{1.0 \linewidth}{!}{
    \begin{tabular}{llccccccccc}
        \toprule
        Modality & Method & Venue & HOTA↑ & DetA↑ & AssA↑ & DetRe↑ & DetPr↑ & AssRe↑ & AssPr↑ & LocA↑ \\
        \midrule
        L-RGB & TransRMOT \cite{wu2023referring} & CVPR 2023 & 1.16 & 0.24 & 5.79 & 0.26 & 2.57 & 5.90 & \textbf{90.28} & 75.00 \\
        L-RGB & TempRMOT \cite{zhang2024bootstrapping} & ArXiv 2024 & 2.75 & 0.72 & 10.58 & 0.82 & 5.28 & 11.94 & 63.42 & 75.07 \\
        L-RGB & YOLOX  \cite{ge2021yolox}+ByteTrack \cite{zhang2022bytetrack}+\uline{iKUN} \cite{du2024ikun} & CVPR 2024 & 7.89 & 2.87 & 21.78 & 4.70 & 6.78 & 25.48 & 62.27 & 77.42 \\
        L-RGB & YOLOX \cite{ge2021yolox}+OC-SORT \cite{cao2023observation}+\uline{iKUN} \cite{du2024ikun} & CVPR 2024 & 8.36 & 3.00 & 23.49 & 4.66 & 7.63 & 27.30 & 65.99 & \textbf{78.32} \\
        L-RGB & CRTracker \cite{chen2025cross} & AAAI 2025 & 8.23 & 2.67 & 27.25 & 3.85 & 7.79 & 29.39 & 66.16 & 72.63 \\
        L-RGB & DKGTrack \cite{li2025language} & ICCV 2025 & 2.99 & 0.42 & 21.71 & 0.44 & 12.71 & 22.28 & 75.55 & 75.46 \\
        L-RGB & \uline{Qwen2.5-VL-3B} \cite{bai2025qwen2}+SORT \cite{bewley2016simple} & ArXiv 2025 & 11.45 & 8.26 & 20.14 & 11.69 & 20.08 & 22.22 & 56.48 & 71.51 \\
        \midrule
        L-RGB-D & \uline{DepthMOT} \cite{wu2024depthmot}+iKUN \cite{du2024ikun} & ArXiv 2024 & 2.41 & 0.88 & 7.67 & 2.68 & 1.27 & 8.62 & 51.08 & 65.72 \\
        L-RGB-D & \uline{DepTR-MOT} \cite{deng2025deptr}+iKUN \cite{du2024ikun} & ArXiv 2025 & 9.51 & 4.83 & 19.26 & 10.86 & 7.69 & 22.68 & 55.65 & 69.15 \\
        L-RGB-D & \uline{SparseTrack} \cite{liu2025sparsetrack}+iKUN \cite{du2024ikun} & TCSVT 2025 & 7.36 & 1.68 & 33.60 & 10.18 & 1.96 & 41.75 & 57.84 & 74.41 \\
        L-RGB-D & \uline{Qwen2.5-VL-3B} \cite{bai2025qwen2}+SORT \cite{bewley2016simple} & ArXiv 2025 & 11.23 & 7.82 & 20.26 & 10.88 & 19.92 & 22.46 & 56.88 & 71.68 \\
        \rowcolor[HTML]{e2f0ff} L-RGB-D & \textbf{DRTrack (Ours)} & - & \textbf{28.25} & \textbf{22.87} & \textbf{35.90} & \textbf{34.53} & \textbf{34.50} & \textbf{49.88} & 45.68 & 71.81 \\
        \bottomrule
    \end{tabular}
}
\end{table*}

\subsection{Evaluation Metrics}

We follow the standard RMOT evaluation protocol \cite{wu2023referring}. HOTA serves as the primary metric because it jointly measures detection and association quality. We additionally report Detection Accuracy (DetA), Association Accuracy (AssA), Detection Recall (DetRe), Detection Precision (DetPr), Association Recall (AssRe), Association Precision (AssPr), and Localization Accuracy (LocA), which together provide complementary evidence on target retrieval, identity preservation, and bounding-box quality.

Because the MLLM dominates the computational cost of the pipeline, we further introduce the \textbf{Accuracy--Efficiency Score (AES)} to quantify the trade-off between tracking accuracy and inference speed. It is defined as the weighted geometric mean of HOTA and the end-to-end throughput:
\begin{equation}
\label{eq:ae_score}
\text{AES} = \mathrm{HOTA}^{\,3/4} \cdot \mathrm{FPS}^{\,1/4}
\end{equation}
where FPS denotes the number of frames processed per second by the complete pipeline, including both MLLM generation and trajectory association, measured on the same GPU. The 3:1 weighting reflects that tracking accuracy is the primary objective of DRMOT, while inference speed is a secondary deployment constraint. As a weighted geometric mean, the AES penalizes configurations that sacrifice either tracking accuracy or inference efficiency disproportionately.

\subsection{Main Results}

Table \ref{tab:main_results} reports the comparison on the DRSet test set. DRTrack achieves a HOTA of \textbf{28.25\%}, which exceeds the strongest compared method, the zero-shot Qwen2.5-VL-3B with SORT under L+RGB inputs, by 16.80 percentage points and more than doubles its score. The improvement holds for both detection and association, as DRTrack obtains the highest DetA of \textbf{22.87\%} and the highest AssA of \textbf{35.90\%}. We analyze the compared methods in three groups below, since each group reveals a different reason why DRMOT cannot be solved by directly applying existing trackers.

The first group consists of RGB-only RMOT trackers. These methods obtain extremely low DetRe, ranging from 0.26\% for TransRMOT to 4.70\% for the YOLOX-based iKUN variants, although several of them keep moderate or high LocA. This pattern indicates that they localize a small number of matched boxes accurately but fail to retrieve most referred targets. TransRMOT illustrates this failure mode most clearly: it reaches the highest AssPr of 90.28\%, yet its HOTA is only 1.16\% because almost all referred targets are missed.  Overall, limited target retrieval is the dominant bottleneck for this group: high LocA or AssPr on the small subset of recovered targets does not translate into strong end-to-end tracking performance.

The second group combines RGB-D multi-object trackers with iKUN, which tests whether depth at the tracking level is sufficient. The answer is negative. SparseTrack with iKUN obtains an AssA of 33.60\%, which approaches that of DRTrack and indicates that depth-aware association helps preserve the identities of generic tracks. However, its DetPr is only 1.96\%, because the selection of referred tracks still relies on RGB-language matching and therefore retains many tracks that do not satisfy the expression. The best method in this group, DepTR-MOT with iKUN, reaches only 9.51\% HOTA. Therefore, depth improves identity preservation, but it cannot compensate for the lack of language grounding.

The third group uses the MLLM as a zero-shot detector. Under SORT, adding the depth image to the input of Qwen2.5-VL-3B changes HOTA from 11.45\% to 11.23\%, so the raw depth image alone does not improve performance. In contrast, DRTrack builds on the same 3B backbone and the same depth input but reaches \textbf{28.25\%}. This contrast is consistent with the analysis in Section \ref{sec:Introduction}: depth becomes effective only when it is incorporated into both language-guided detection and cross-frame association.  The AssPr of DRTrack (45.68\%) is lower than that of the zero-shot MLLM baselines, which is likely because DRTrack retains many more referred trajectories and thus faces more difficult identity decisions. Its LocA (71.81\%) is also lower than that of the detector-based iKUN variants (up to 78.32\%), which suggests that the coordinate regression of the MLLM remains less precise than that of a dedicated detector.

\begin{table*}[t]
\centering
\caption{Comparison of different Multimodal Large Language Models (MLLMs) on the test set of the DRSet dataset under the L+RGB input setting. ↑ indicates that higher score is better. The best results are marked in \textbf{bold}.}
\label{tab:mllm_backbone}
\footnotesize
\begin{tabularx}{\linewidth}{l*{11}{>{\centering\arraybackslash}X}}
    \toprule
    \multicolumn{2}{c}{MLLMs} & \multicolumn{10}{c}{Metrics} \\
    \cmidrule(lr){1-2} \cmidrule(lr){3-12}
    Name & Size & HOTA↑ & DetA↑ & AssA↑ & DetRe↑ & DetPr↑ & AssRe↑ & AssPr↑ & LocA↑ & FPS↑ & AES↑ \\
    \midrule
    DeepSeek-VL \cite{lu2024deepseek} & 7B & 9.76 & 9.79 & 10.53 & 19.49 & 14.32 & 12.43 & 39.65 & 63.46 & 5.63 & 8.51 \\
    InternVL3.5 \cite{wang2025internvl3} & 8B & 5.61 & 6.49 & 5.29 & 9.20 & 15.82 & 5.78 & 49.86 & 66.28 & 2.34 & 4.51 \\
    LLaVA-NEXT \cite{liu2024llavanext} & 8B & 14.37 & 14.24 & 16.26 & 27.05 & 20.67 & 18.92 & 50.11 & 69.75 & 4.71 & 10.87 \\
    Qwen2.5-VL \cite{bai2025qwen2} & 3B & 11.45 & 8.26 & \textbf{20.14} & 11.69 & 20.08 & \textbf{22.22} & \textbf{56.48} & 71.51 & \textbf{5.83} & 9.67 \\
    \midrule
    \rowcolor[HTML]{e2f0ff} Qwen2.5-VL \cite{bai2025qwen2} & 7B & \textbf{17.70} & \textbf{18.76} & 18.06 & \textbf{41.71} & \textbf{23.23} & 21.13 & 50.02 & \textbf{72.76} & 3.75 & \textbf{12.01} \\
    \bottomrule
\end{tabularx}
\end{table*}

\subsection{Backbone Analysis}

Before analyzing the components of DRTrack, we justify the choice of the MLLM backbone. Table \ref{tab:mllm_backbone} compares representative MLLMs as zero-shot detectors under the L+RGB setting with SORT association. Qwen2.5-VL-7B achieves the best HOTA of \textbf{17.70\%} and the highest DetRe of 41.71\%, which shows that a larger model retrieves more candidate targets. However, its AssA of 18.06\% is lower than the 20.14\% of Qwen2.5-VL-3B, so stronger frame-level perception does not by itself produce temporally stable tracks. Qwen2.5-VL-3B obtains the best AssA, AssRe, and AssPr among all evaluated models and the highest throughput of 5.83 FPS, which is 1.55 times that of the 7B model. Although the 7B model attains a higher zero-shot AES, the 3B model substantially reduces the cost of GRPO training and online inference, and we therefore adopt it as the default backbone.

This choice also allows us to test whether the proposed adaptation can compensate for a smaller model. Built on the 3B backbone, DRTrack exceeds the zero-shot 7B model by 10.55 percentage points in HOTA. Compared with this model, DRTrack raises DetPr from 23.23\% to \textbf{34.50\%} and AssRe from 21.13\% to \textbf{49.88\%}. Therefore, the improvement of DRTrack does not come from model scale, and the following ablations identify the components responsible for it.

\begin{table*}[t]
\centering
\caption{Component ablation of DRTrack with Qwen2.5-VL-3B-Instruct \cite{bai2025qwen2} on the test set of the DRSet dataset. ``Ours'' denotes the proposed depth-enhanced association. ↑ indicates that higher score is better. The best results are marked in \textbf{bold}.}
\label{tab:component_study}
\footnotesize
\begin{tabularx}{\linewidth}{l*{10}{>{\centering\arraybackslash}X}}
    \toprule
    \multicolumn{3}{c}{Settings} & \multicolumn{8}{c}{Metrics} \\
    \cmidrule(lr){1-3} \cmidrule(lr){4-11}
    Input Data & Association & GRPO & HOTA↑ & DetA↑ & AssA↑ & DetRe↑ & DetPr↑ & AssRe↑ & AssPr↑ & LocA↑ \\
    \midrule
    L+RGB   & SORT & \ding{55} & 11.45 & 8.26 & 20.14 & 11.69 & 20.08 & 22.22 & 56.48 & 71.51 \\
    L+RGB+D & SORT & \ding{55} & 11.23 & 7.82 & 20.26 & 10.88 & 19.92 & 22.46 & \textbf{56.88} & 71.68 \\
    \midrule
    L+RGB   & Ours & \ding{55} & 13.29 & 8.26 & 24.97 & 10.35 & 25.95 & 29.91 & 51.62 & 71.96 \\
    L+RGB+D & Ours & \ding{55} & 12.75 & 7.70 & 24.66 & 9.58 & 25.40 & 29.80 & 51.90 & \textbf{72.12} \\
    \midrule
    L+RGB   & SORT & \ding{51} & 15.62 & 14.84 & 17.84 & \textbf{45.33} & 16.91 & 21.08 & 47.92 & 72.40 \\
    \rowcolor[HTML]{e2f0ff} L+RGB+D & Ours & \ding{51} & \textbf{28.25} & \textbf{22.87} & \textbf{35.90} & 34.53 & \textbf{34.50} & \textbf{49.88} & 45.68 & 71.81 \\
    \bottomrule
\end{tabularx}
\end{table*}

\begin{table}[t]
\centering
\caption{Ablation of the reward functions in Geometric-Aware GRPO. All variants use Qwen2.5-VL-3B-Instruct \cite{bai2025qwen2}, L+RGB+D inputs, and the proposed association on the test set of the DRSet dataset. ↑ indicates that higher score is better. The best results are marked in \textbf{bold}.}
\label{tab:reward_ablation}
\resizebox{1.0 \linewidth}{!}{
    \setlength{\tabcolsep}{1.0mm}{
        \begin{tabular}{cccccccccc}
            \toprule
            \multicolumn{2}{c}{Reward} & \multicolumn{8}{c}{Metrics} \\
            \cmidrule(lr){1-2} \cmidrule(lr){3-10}
            $R_{\text{format}}$ & $R_{\text{IoU}}$ & HOTA↑ & DetA↑ & AssA↑ & DetRe↑ & DetPr↑ & AssRe↑ & AssPr↑ & LocA↑ \\
            \midrule
            \ding{51} & \ding{55} & 27.28 & 22.01 & 34.78 & 34.23 & 32.84 & 48.24 & 44.95 & 71.79 \\
            \ding{55} & \ding{51} & 27.10 & 21.94 & 34.46 & \textbf{35.53} & 31.56 & 47.25 & 45.13 & 71.73 \\
            \rowcolor[HTML]{e2f0ff} \ding{51} & \ding{51} & \textbf{28.25} & \textbf{22.87} & \textbf{35.90} & 34.53 & \textbf{34.50} & \textbf{49.88} & \textbf{45.68} & \textbf{71.81} \\
            \bottomrule
        \end{tabular}
    }
}
\end{table}

\begin{table*}[t]
\centering
\caption{Analysis of fusion weights $\alpha$ in Equation (\ref{eq:RGBD joint similarity.}). ↑ indicates that higher score is better. The best results are marked in \textbf{bold}.}
\label{tab:component_ablation}
\footnotesize
\begin{tabularx}{\linewidth}{lccl*{9}{>{\centering\arraybackslash}X}}
    \toprule
    \multicolumn{3}{c}{Multimodal Large Language Models (MLLMs)} & \multirow{2}{*}{Input Data} & Fusion Weight & \multicolumn{8}{c}{Metrics} \\
    \cmidrule(lr){1-3} \cmidrule(lr){5-5} \cmidrule(lr){6-13}
    Name & Size & RL Fine-tuning & & $\alpha$ & HOTA↑ & DetA↑ & AssA↑ & DetRe↑ & DetPr↑ & AssRe↑ & AssPr↑ & LocA↑ \\
    \midrule
    Qwen2.5-VL-3B-Instruct & 3B & \ding{51} (GRPO) & L+RGB+D & 0.0 & 28.18 & 22.80 & 35.85 & 34.47 & 34.39 & \textbf{49.95} & 45.31 & 71.79 \\
    Qwen2.5-VL-3B-Instruct & 3B & \ding{51} (GRPO) & L+RGB+D & 0.1 & 28.24 & 22.86 & 35.89 & 34.53 & 34.47 & 49.88 & 45.67 & \textbf{71.81} \\
    Qwen2.5-VL-3B-Instruct & 3B & \ding{51} (GRPO) & L+RGB+D & 0.2 & 28.23 & 22.87 & 35.87 & 34.54 & 34.49 & 49.84 & 45.65 & \textbf{71.81} \\
    Qwen2.5-VL-3B-Instruct & 3B & \ding{51} (GRPO) & L+RGB+D & 0.4 & 28.21 & 22.82 & 35.89 & 34.53 & 34.38 & 49.88 & 45.68 & 71.81 \\
    Qwen2.5-VL-3B-Instruct & 3B & \ding{51} (GRPO) & L+RGB+D & 0.5 & 28.10 & 22.67 & 35.84 & 34.56 & 34.03 & 49.76 & 45.72 & 71.81 \\
    Qwen2.5-VL-3B-Instruct & 3B & \ding{51} (GRPO) & L+RGB+D & 0.6 & 27.60 & 22.26 & 35.24 & 34.65 & 33.02 & 48.21 & 46.29 & 71.80 \\
    Qwen2.5-VL-3B-Instruct & 3B & \ding{51} (GRPO) & L+RGB+D & 0.7 & 19.87 & 18.50 & 22.64 & 37.13 & 24.17 & 27.19 & 49.58 & 71.59 \\
    Qwen2.5-VL-3B-Instruct & 3B & \ding{51} (GRPO) & L+RGB+D & 0.8 & 17.82 & 17.57 & 19.43 & 37.89 & 22.33 & 22.94 & \textbf{49.74} & 71.55 \\
    Qwen2.5-VL-3B-Instruct & 3B & \ding{51} (GRPO) & L+RGB+D & 0.9 & 16.93 & 16.98 & 18.46 & 38.33 & 21.24 & 21.52 & 49.60 & 71.50 \\
    Qwen2.5-VL-3B-Instruct & 3B & \ding{51} (GRPO) & L+RGB+D & 1.0 & 16.11 & 16.64 & 17.22 & \textbf{38.69} & 20.60 & 19.95 & 49.38 & 71.40 \\
    \midrule
    \rowcolor[HTML]{e2f0ff} Qwen2.5-VL-3B-Instruct & 3B & \ding{51} (GRPO) & L+RGB+D & 0.3 & \textbf{28.25} & \textbf{22.87} & \textbf{35.90} & 34.53 & \textbf{34.50} & 49.88 & 45.68 & 71.81 \\
    \bottomrule
\end{tabularx}
\end{table*}

\subsection{Ablation Study}

Table \ref{tab:component_study} examines how the input modality, the association strategy, and the GRPO-adapted detection stage contribute to the final performance. We first consider the non-finetuned detector. Feeding the depth image directly into the MLLM does not improve tracking, as HOTA changes from 11.45\% to 11.23\% under SORT and from 13.29\% to 12.75\% under the proposed association. This result confirms that depth cannot be treated as an unprocessed auxiliary image. Replacing SORT with the proposed association improves HOTA from 11.45\% to 13.29\% under L+RGB inputs and from 11.23\% to 12.75\% under L+RGB+D inputs, mainly through higher AssA and AssRe. However, DetRe remains around 10\% in all four settings, which shows that association cannot recover targets that the detector fails to retrieve. The gain from association is therefore bounded by the recall of the detection stage.

The GRPO-adapted detection stage removes this bound. With L+RGB+D inputs and the proposed association, it raises HOTA from 12.75\% to \textbf{28.25\%}, DetRe from 9.58\% to \textbf{34.53\%}, and DetPr from 25.40\% to \textbf{34.50\%}. As described in the experimental setup, this comparison covers the complete adapted detection stage, including its RGB-D instruction. The resulting detections also benefit association, as AssRe increases from 29.80\% to \textbf{49.88\%}. The importance of pairing better detection with depth-enhanced association is shown by the L+RGB detector adapted with GRPO and combined with SORT. This variant achieves the highest DetRe of 45.33\%, but its DetPr is only 16.91\% and its AssA drops to 17.84\%, which limits its HOTA to 15.62\%. Higher recall does not translate into reliable trajectories unless the association stage can maintain the identities of the detections. These results support the staged design motivated in Section \ref{sec:Introduction}: the detection stage determines which targets are retrieved, and the association stage determines whether their identities are preserved.

Table \ref{tab:reward_ablation} analyzes the two rewards used in GRPO. The combined format and IoU rewards achieve the highest HOTA of 28.25\% and DetPr of 34.50\% among the evaluated reward configurations. The two rewards affect detection differently. Without the format reward, DetRe increases to 35.53\% while DetPr decreases to 31.56\%, which indicates that the model produces more boxes but fewer correct ones. Without the IoU reward, DetPr decreases to 32.84\% because the policy no longer receives direct feedback on localization quality. The full reward achieves the best balance between recall and precision. We further examine the amount of GRPO training data. Using the full training set instead of $10\%$ of it raises HOTA from 28.25\% to only 28.77\%, an absolute gain of 0.52 percentage points. Therefore, we keep the $10\%$ setting as the default, which shows that DRTrack achieves strong results with limited task-specific fine-tuning.

Table \ref{tab:component_ablation} isolates the contribution of depth within the association stage by varying the fusion weight $\alpha$ in Equation (\ref{eq:RGBD joint similarity.}). When $\alpha=1.0$, the similarity reduces to 2D IoU and HOTA falls to 16.11\%, with AssA dropping to 17.22\% and AssRe to 19.95\%. This score is close to the 15.62\% of the IoU-based SORT variant in Table \ref{tab:component_study}, which confirms that most of the association gain comes from the depth similarity term. HOTA remains between 28.10\% and 28.25\% for $\alpha$ from 0.0 to 0.5, decreases slightly to 27.60\% at $\alpha=0.6$, and drops sharply to 19.87\% at $\alpha=0.7$. As $\alpha$ increases beyond this point, DetRe rises while DetPr falls. This trend suggests that weaker depth constraints produce more fragmented tracks, whose propagated boxes increase recall at the cost of precision. Using depth alone ($\alpha=0.0$) yields 28.18\%, slightly below the 28.25\% obtained with $\alpha=0.3$, so 2D overlap serves as a complementary cue with a small but consistent benefit. These results verify that depth provides the cross-frame geometric consistency required to resolve association ambiguity.

\begin{table}[t]
\centering
\caption{Sensitivity to the depth scaling factor $\sigma$ on the DRSet test split. ↑ indicates that higher score is better. The best results are marked in \textbf{bold}.}
\label{tab:sigma_sensitivity}
\footnotesize
\setlength{\tabcolsep}{1.5pt}
\begin{tabularx}{\linewidth}{c*{8}{>{\centering\arraybackslash}X}}
    \toprule
    \multirow{2}{*}{$\sigma$} & \multicolumn{8}{c}{Metrics} \\
    \cmidrule(lr){2-9}
    & HOTA↑ & DetA↑ & AssA↑ & DetRe↑ & DetPr↑ & AssRe↑ & AssPr↑ & LocA↑ \\
    \midrule
    1 & 20.54 & 19.31 & 23.13 & \textbf{35.73} & 26.29 & 29.34 & \textbf{46.99} & 71.66 \\
    5 & 26.95 & 22.00 & 34.05 & 34.84 & 32.29 & 46.18 & 46.41 & 71.79 \\
    10 & 27.98 & 22.54 & 35.75 & 34.60 & 33.69 & 49.51 & 45.83 & 71.80 \\
    15 & 28.17 & 22.78 & 35.84 & 34.54 & 34.29 & 49.79 & 45.66 & \textbf{71.82} \\
    30 & 28.24 & 22.87 & \textbf{35.90} & 34.53 & 34.50 & \textbf{49.88} & 45.68 & 71.81 \\
    35 & 28.24 & 22.87 & 35.89 & 34.53 & \textbf{34.51} & \textbf{49.88} & 45.68 & 71.81 \\
    50 & 28.24 & \textbf{22.88} & 35.89 & 34.53 & \textbf{34.51} & 49.87 & 45.68 & 71.81 \\
    \midrule
    \rowcolor[HTML]{e2f0ff} 20 & \textbf{28.25} & 22.87 & \textbf{35.90} & 34.53 & 34.50 & \textbf{49.88} & 45.68 & 71.81 \\
    \bottomrule
\end{tabularx}
\end{table}

\begin{table}[t]
\centering
\caption{Robustness to depth noise on the DRSet test split. ``Ours'' denotes the original depth input without perturbation. ↑ indicates that higher score is better. The best results are marked in \textbf{bold}.}
\label{tab:depth_noise}
\footnotesize
\begin{tabularx}{\linewidth}{l*{4}{>{\centering\arraybackslash}X}}
    \toprule
    \multirow{2}{*}{Depth Setting} & \multicolumn{4}{c}{Metrics} \\
    \cmidrule(lr){2-5}
    & HOTA↑ & DetA↑ & AssA↑ & AssRe↑ \\
    \midrule
    Gaussian 20 mm       & 28.23 & \textbf{22.88} & 35.83 & 49.83 \\
    Gaussian 50 mm       & 28.24 & \textbf{22.88} & 35.84 & 49.81 \\
    Gaussian 100 mm      & 28.22 & 22.87 & 35.80 & 49.80 \\
    Multiplicative 5\%   & 28.23 & 22.87 & 35.81 & 49.80 \\
    Multiplicative 10\%  & 28.23 & \textbf{22.88} & 35.85 & 49.87 \\
    Dropout 5\%          & 28.09 & 22.69 & 35.76 & 49.72 \\
    Dropout 10\%         & 28.05 & 22.66 & 35.66 & 49.66 \\
    Block drop           & 28.22 & 22.84 & 35.86 & 49.84 \\
    Mixed mild           & 28.11 & 22.76 & 35.69 & 49.70 \\
    Mixed strong         & \textbf{28.25} & 22.84 & \textbf{35.95} & \textbf{50.02} \\
    \midrule
    \rowcolor[HTML]{e2f0ff} Ours & \textbf{28.25} & 22.87 & 35.90 & 49.88 \\
    \bottomrule
\end{tabularx}
\end{table}

\begin{table}[t]
\centering
\caption{Sensitivity to the MLLM inference interval on the DRSet test split. ↑ indicates that higher score is better. The best results are marked in \textbf{bold}.}
\label{tab:inference_interval}
\footnotesize
\begin{tabularx}{\linewidth}{c*{7}{>{\centering\arraybackslash}X}}
    \toprule
    Interval & \multicolumn{7}{c}{Metrics} \\
    \cmidrule(lr){2-8}
    (frames) & HOTA↑ & DetA↑ & AssA↑ & DetRe↑ & AssRe↑ & FPS↑ & AES↑ \\
    \midrule
    1  & \textbf{38.36} & \textbf{32.49} & \textbf{45.89} & \textbf{48.07} & \textbf{60.61} & 0.49 & 12.88 \\
    5  & 33.76 & 27.80 & 41.89 & 40.96 & 56.21 & 2.42 & 17.46 \\
    \rowcolor[HTML]{e2f0ff} 10 & 28.25 & 22.87 & 35.90 & 34.53 & 49.88 & 4.82 & \textbf{18.16} \\
    20 & 21.76 & 16.95 & 29.05 & 26.69 & 42.12 & 9.15 & 17.52 \\
    30 & 19.14 & 14.38 & 26.80 & 23.13 & 39.21 & \textbf{12.94} & 17.36 \\
    \bottomrule
\end{tabularx}
\end{table}

\subsection{Parameter Sensitivity}

Having established the role of depth in association, we next examine how sensitive this role is to the depth scaling factor $\sigma$ and to the quality of the depth input. As shown in Table \ref{tab:sigma_sensitivity}, HOTA increases from 20.54\% at $\sigma=1$ to 28.25\% at $\sigma=20$ and then varies by no more than 0.01 percentage points for $\sigma$ between 20 and 50. A small $\sigma$ makes the depth similarity decay too quickly, so that ordinary depth fluctuations of the same target are penalized as if they indicated different objects. Once $\sigma$ is sufficiently large, the depth term separates targets at clearly different distances without reacting to small variations. Therefore, we use $\sigma=20$ as the default setting.

\begin{figure*}[t]
    \centering
    \includegraphics[width=1.0\linewidth]{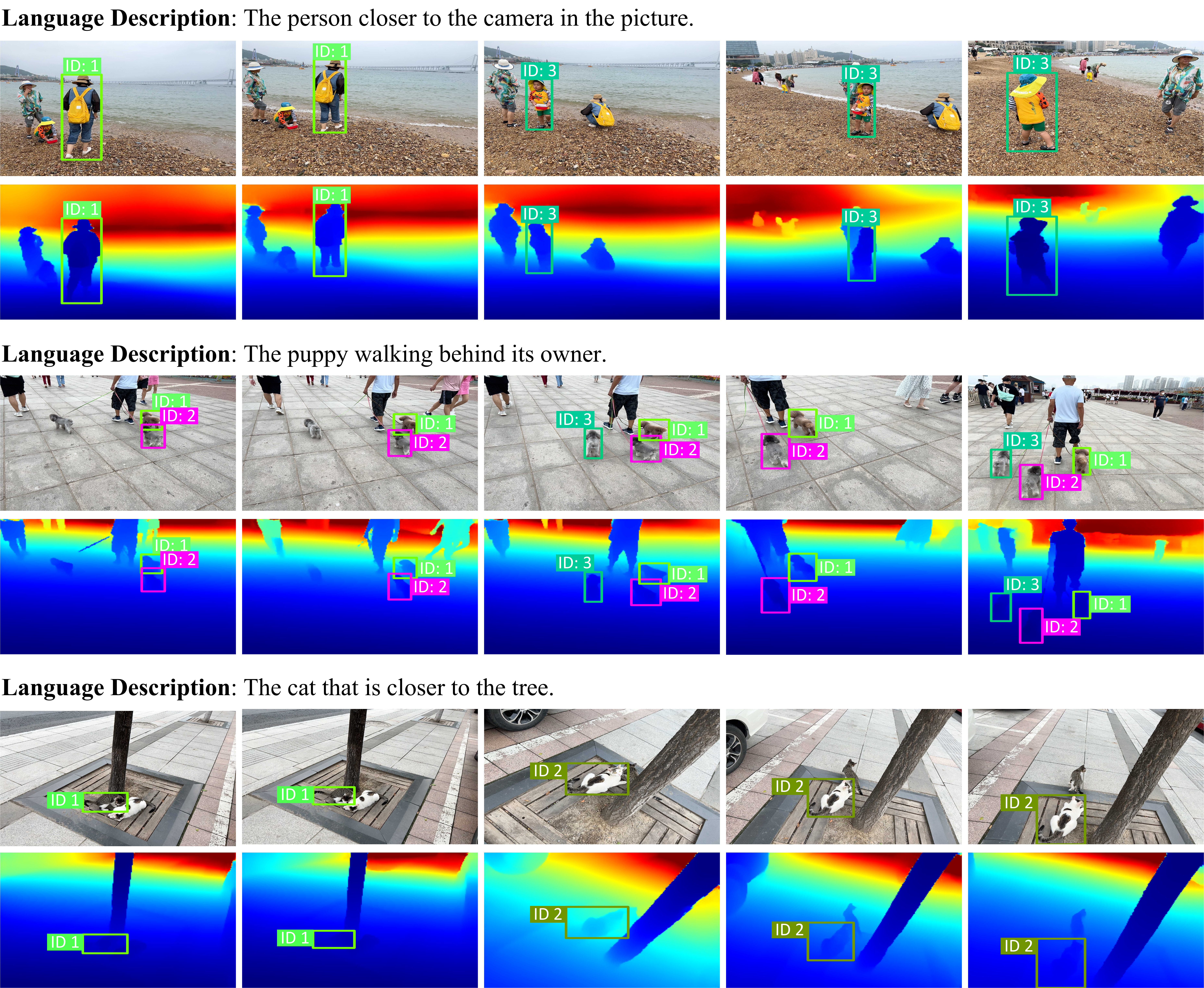}
    \caption{Qualitative results of DRTrack on the DRSet dataset. For each example, the top row shows RGB frames with predicted bounding boxes and identities, while the bottom row shows the corresponding depth maps. The examples illustrate how depth cues help resolve depth-dependent spatial descriptions, such as ``closer to'' and ``behind'', while maintaining temporally consistent identities.}
    \label{fig:Qualitative Results.}
\end{figure*}

Table \ref{tab:depth_noise} evaluates the robustness of DRTrack to controlled depth perturbations. Across all tested settings, HOTA ranges from 28.05\% to 28.25\%. The largest decrease, 0.20 percentage points, occurs under 10\% pixel dropout, whereas Gaussian and multiplicative noise change HOTA by at most 0.03 percentage points. This behavior is consistent with the design of the depth similarity, which averages the depth over all valid pixels inside a box and is therefore insensitive to zero-mean pixel noise, while dropout reduces the number of valid pixels used for this average. Together with the insensitivity to $\sigma$, this result indicates that DRTrack relies on coarse depth differences between targets rather than on precise depth values. However, these experiments do not cover all possible sensor errors or patterns of missing depth.

Finally, Table \ref{tab:inference_interval} studies the MLLM inference interval, which controls the trade-off between accuracy and efficiency. Shorter intervals provide more frequent language-guided detections and trajectory corrections. Invoking the MLLM at every frame yields the best HOTA of 38.36\%, but its throughput of 0.49 FPS is impractical for online tracking. Conversely, the 30-frame interval reaches the highest throughput of 12.94 FPS at a HOTA of 19.14\%. The AES defined in Equation (\ref{eq:ae_score}) quantifies this trade-off and peaks at the 10-frame interval with a value of 18.16. This setting (28.25\%) retains 73.6\% of the best HOTA while requiring only one tenth of the MLLM calls. Notably, DRTrack outperforms all compared methods in Table \ref{tab:main_results} at every tested interval, including the 30-frame setting. Thus, we adopt the 10-frame interval as the default and regard shorter intervals as higher-cost variants for accuracy-critical.

\subsection{Qualitative Results}

Fig. \ref{fig:Qualitative Results.} presents qualitative examples of DRTrack handling depth-dependent referring expressions. In the first example, ``The person closer to the camera in the picture'' perspective effects cause several individuals to share similar 2D visual scales, making appearance cues highly ambiguous. The depth map resolves this by providing an explicit physical distance metric, enabling the model to accurately ground the spatial semantics and localize the correct target.

The framework also demonstrates robustness under severe occlusions and complex relative spatial constraints. In the second example, ``The puppy walking behind its owner'' high bounding-box overlap creates unstable appearance features that typically lead to identity switches. The depth map reveals the ordinal spatial relation between the overlapping targets, allowing the depth-enhanced association to maintain stable identities. Similarly, for ``The cat that is closer to the tree'' explicit depth assists the GRPO-adapted detector in selecting the target that satisfies the relative distance constraint, while providing supplementary geometric evidence for stable cross-frame matching.

These qualitative cases align consistently with our quantitative findings. They demonstrate that depth information resolves geometric ambiguities effectively when integrated via Geometric-Aware GRPO fine-tuning and RGBD joint association, rather than as mere visual input.

\section{Conclusion}
\label{sec:Conclusion}

In this work, we propose a novel task, RGB\textbf{D} \textbf{R}eferring \textbf{M}ulti-\textbf{O}bject \textbf{T}racking (\textbf{DRMOT}). This task aims to address key limitations of existing 2D referring multi-object tracking methods in handling complex spatial ambiguities and occlusions. To advance research on the DRMOT task, we construct \textbf{DRSet}, a tailored dataset designed to evaluate models’ spatial-semantic detection and tracking capabilities in realistic 3D perception conditions. Furthermore, we propose \textbf{DRTrack}, an innovative tracking framework that tightly couples an MLLM, fine-tuned via Geometric-Aware GRPO, with a Depth-Enhanced Association mechanism. By incorporating explicit depth cues and a novel RGBD joint similarity metric, DRTrack effectively eliminates 2D spatial ambiguities and maintains highly robust trajectory associations. Extensive experiments demonstrate that DRTrack achieves state-of-the-art (SOTA) performance, establishing a baseline for the DRMOT task.

{\small
\bibliographystyle{IEEEtran}
\bibliography{reference}
}

\vfill

\end{document}